\documentclass[10pt,twocolumn,letterpaper]{article}

\usepackage{iccv}
\usepackage{times}
\usepackage{epsfig}
\usepackage{graphicx}
\usepackage{amsmath}
\usepackage{amssymb}

\usepackage{makecell}
\usepackage{enumitem}
\usepackage{booktabs}

\usepackage{subfigure}
\usepackage{algorithm}
\usepackage{algorithmic}
\newcommand{\xdownarrow}[1]{%
  {\left\downarrow\vbox to #1{}\right.\kern-\nulldelimiterspace}
}
\usepackage{nccmath}

\usepackage{multirow}

\usepackage[pagebackref=true,breaklinks=true,letterpaper=true,colorlinks,bookmarks=false]{hyperref}

\iccvfinalcopy 

\ificcvfinal\fi

\begin{document}

\title{Noise Injection-based Regularization for Point Cloud Processing}

\author{Xiao Zang\thanks{Department of Electrical and Computer Engineering, Rutgers University. Correspondence to: bo.yuan@soe.rutgers.edu}\\
\and
Yi Xie\footnotemark[1]\\
\and 
Siyu Liao\footnotemark[1]\\
\and 
Jie Chen\thanks{MIT-IBM Watson AI Lab, IBM Research.}\\
\and Bo Yuan\footnotemark[1]
}


\maketitle
\ificcvfinal\thispagestyle{empty}\fi

\begin{abstract}
Noise injection-based regularization, such as Dropout, has been widely used in image domain to improve the performance of deep neural networks (DNNs). However, efficient regularization in the point cloud domain is rarely exploited, and most of the state-of-the-art works focus on data augmentation-based regularization. In this paper, we, for the first time, perform systematic investigation on noise injection-based regularization for point cloud-domain DNNs. To be specific, we propose a series of regularization techniques, namely DropFeat, DropPoint and DropCluster, to perform noise injection on the point feature maps at the feature level, point level and cluster level, respectively. We also empirically analyze the impacts of different factors, including dropping rate, cluster size and dropping position, to obtain useful insights and general deployment guidelines, which can facilitate the adoption of our approaches across different datasets and DNN architectures.

We evaluate our proposed approaches on various DNN models for different point cloud processing tasks. Experimental results show our approaches enable significant performance improvement. Notably, our DropCluster brings $1.5\%$, $1.3\%$ and $0.8\%$ higher overall accuracy for PointNet, PointNet++ and DGCNN, respectively, on ModelNet40 shape classification dataset. On ShapeNet part segmentation dataset, DropCluster brings $0.5\%$, $0.5\%$ and $0.2\%$ mean Intersection-over-union (IoU) increase for PointNet, PointNet++ and DGCNN, respectively. On S3DIS semantic segmentation dataset, DropCluster improves the mean IoU of PointNet, PointNet++ and DGCNN by $3.2\%$, $2.9\%$ and $3.7\%$, respectively. Meanwhile, DropCluster also enables the overall accuracy increase for these three popular backbone DNNs by $2.4\%$, $2.2\%$ and $1.8\%$, respectively.

\end{abstract}

\section{Introduction}

\label{sec:intro}
Point cloud, as the commonly used 3D data representation, has been widely generated, collected and processed in many important computer vision applications, such as autonomous driving, AR/VR, remote sensing etc. Motivated by the current unprecedented success and popularity of deep neural networks (DNNs) in the 2D image processing, both academia and industry are now actively investigating the potential high-performance DNN-based solutions for the efficient 3D point cloud processing.

However, different from the spatially-regular image, point cloud is essentially the unordered set of vectors, which are inherently invariant to the permutation of the member points. Such important and unique characteristics of point cloud data, if not properly considered, would significantly limit the effectiveness of DNNs. To address this challenge, some earlier efforts~\cite{wu20153d, maturana2015voxnet} propose to first convert the irregular point cloud to the regular volumetric representation, and then use the standard 3D convolutional neural networks (CNNs) for backend processing. Such intermediate representation-based strategy, though facilitating the convenient utilization of the existing DNN models, suffers high demand on memory usage as well as inevitable quantization artifacts. Therefore, pioneered by PointNet~\cite{qi2017pointnet}, directly consuming the raw point cloud data has become the preferred solution. To date, many different DNN architectures (e.g., PointNet~\cite{qi2017pointnet}, PointNet++~\cite{qi2017pointnet}, DGCNN~\cite{wang2019dynamic}) have been proposed to process 3D point clouds without extra voxelization or data transformation, and some of them demonstrate the state-of-the-art performance in various point cloud processing tasks.

\begin{figure*}
    \centering
    \includegraphics[width=\textwidth]{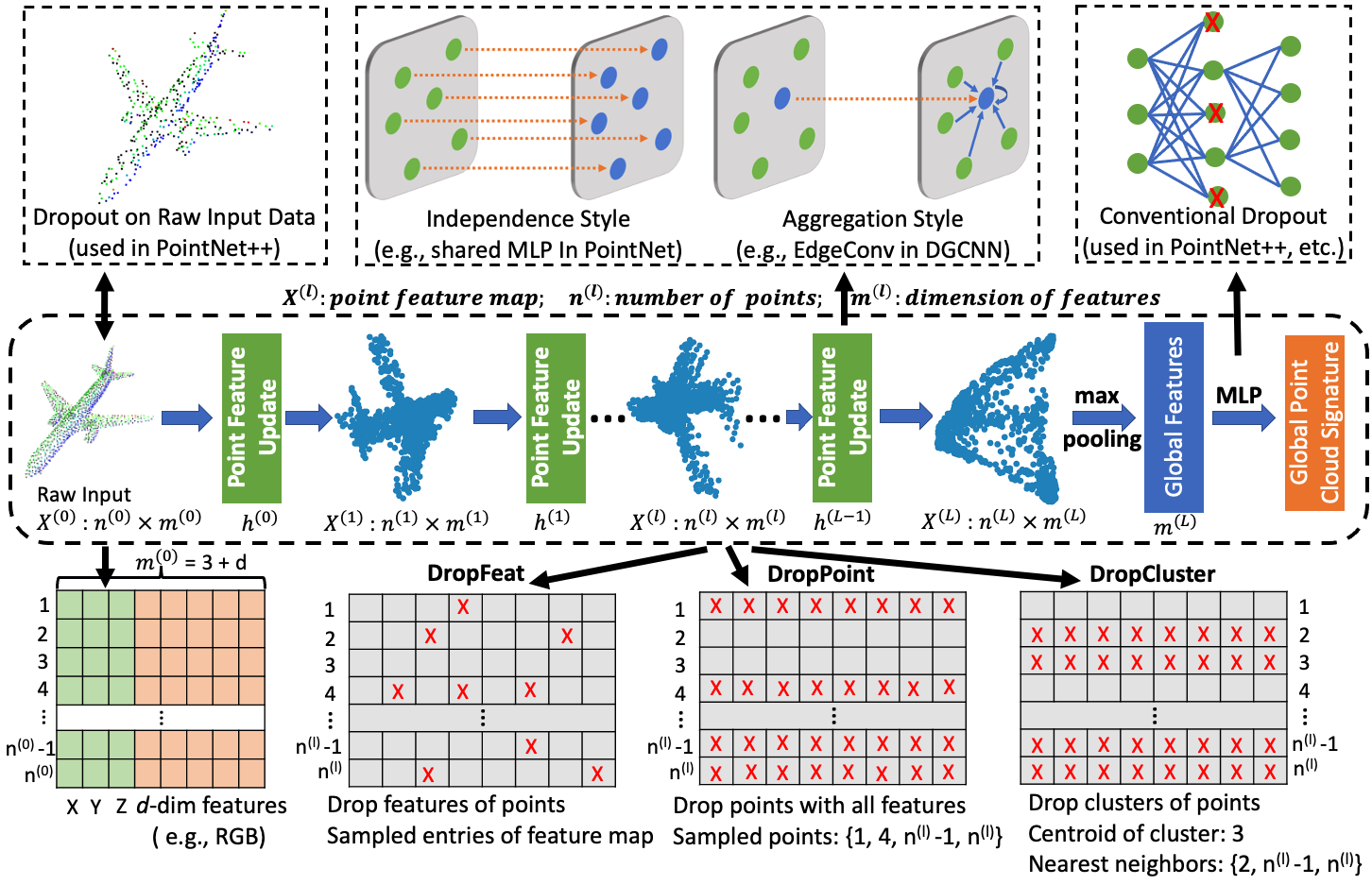}
    \caption{Our proposed noise injection-based regularization (DropFeat, DropPoint and DropCluster) for point cloud processing.}
    \label{fig:framework}
    \vspace{-5mm}
\end{figure*}

    

\textbf{Regularization for Point Cloud: Importance \& Benefits.} These recent architecture-level innovations in the point cloud domain indeed show the huge benefits brought by the advancement of DNN architecture design. However, on the other aspect, \textit{regularization}, as another important design strategy that has significantly promoted the development of deep learning in the image domain, is little exploited in the DNN-based point cloud processing. In principle, regularization, if performed properly, can potentially provide even more significant performance improvement in the point cloud domain than what it has done in the image domain. As pointed out in~\cite{lee2021regularization}, 3D point cloud datasets are usually much smaller and less diverse than 2D image datasets (e.g., 40-category 10K-model ModelNet40 vs 1000-class 1000K-image ImageNet). Therefore, compared with their counterparts for image processing, the DNN models for point cloud processing are typically more prone to overfitting and lacking generality. These severe challenges, fortunately, are just what regularization can effectively alleviate.

\textbf{Noise Injection-based Regularization for Point Cloud: Uncharted Territory.} Despite its promising potential benefits, regularization has not been thoroughly studied in the point cloud domain. To date, only very few works exploit this direction, and most of them~\cite{lee2021regularization, 10.1007/978-3-030-58580-8_20} focus on \textit{data augmentation-based} regularization, which essentially performs manipulation on the \textit{raw input points}. On the other side, \textit{noise injection-based} regularization, e.g., Dropout, DropBlock and DropPath~\cite{srivastava2014dropout, ghiasi2018dropblock, cai2019effective}, as an important regularization strategy that has been widely adopted in the image-domain DNNs, is very little studied in the point cloud processing. Currently only the conventional Dropout~\cite{srivastava2014dropout}, which randomly drops part of the 1D activation outputs, is straightforwardly used on the back-end last multilayer perceptron (MLP) of the point cloud-domain DNNs. Surprisingly, how to properly drop the information-rich high-dimensional \textit{point features} generated by the front-end point processing layers (e.g., shared MLPs in PointNet and EdgeConv in DGCNN), which is potentially much more important and critical to alleviate overfitting problem and improve model generality, is not investigated and reported in the existing literature.

\textbf{Technical Preview and Contributions.} In this paper, we, for the first time, systematically analyze and investigate the noise injection-based regularization for point cloud processing, and propose a series of regularization techniques for the point cloud-domain DNNs. To be specific, we propose three different regularization solutions, namely \textbf{DropFeat}, \textbf{DropPoint} and \textbf{DropCluster}, to drop information of the point features of DNN at feature level, point level, and cluster level, respectively. Similar to their counterparts in the image domain, these point cloud-domain noise injection approaches are easy for implementation. More importantly, they can be used as convenient plug-ins to improve various backbone DNN architectures for various classification/segmentation tasks. In overall, the contributions of this paper are summarized as follows:
\begin{itemize}
    \item We systematically investigate the noise injection-based regularization for point cloud processing, and propose to regularize point cloud-domain DNN models at three different levels (feature, point and cluster). Our proposed three regularization solutions (DropFeat, DropPoint and DropCluster) are easy for implementation, and they are very general and effective for different DNN models on different datasets. To the best of our knowledge, this is the first work that comprehensively and systematically studies the noise injection-based regularization in the point cloud domain.
    \item We empirically analyze the impacts of different dropping factors on the regularization performance. Based on our ablation study for the dropping rate, the cluster size and the dropping positions, we obtain useful insights and general guidelines that facilitate the deployment of our regularization techniques across different datasets and different DNN architectures.  
    \item We perform extensive experiments for various DNN models in various tasks. on ModelNet40 shape classification dataset, DropCluster enables $1.5\%$, $1.3\%$ and $0.8\%$ overall accuracy increase for PointNet, PointNet++ and DGCNN, respectively. In part segmentation task, DropCluster brings $0.5\%$, $0.5\%$ and $0.2\%$ mean IoU increase for PointNet, PointNet++ and DGCNN, respectively, on ShapeNet dataset. On S3DIS semantic segmentation dataset, DropCluster increases the mean IoU of PointNet, PointNet++ and DGCNN by $3.2\%$, $2.9\%$ and $3.7\%$, respectively. Also, the overall accuracy for these three models are improved by $2.4\%$, $2.2\%$ and $1.8\%$, respectively.
\end{itemize}

\section{Related Work} 
\label{sec:related}
\textbf{Deep Learning on Point Clouds.} The conventional DNN models are designed to process input data with regular structure. To adapt deep learning for the irregular point cloud data-based tasks, a simple solution is to voxelize the point cloud to a volumetric representation, which can be then processed by various well established 3D CNNs~\cite{liu2019point, zhou2018voxelnet, ben20173d}. A major drawback of this intermediate representation-based strategy is the high memory cost incurred by the voxelization. Meanwhile, this approach is also limited by the inevitable quantization artifact. Multiview-based solutions~\cite{su2015multi, yu2018multi, yang2019learning, qi2016volumetric, wang2019dominant} render and project the 3D point clouds to multiple 2D images, and then apply the well engineered 2D CNNs to perform the corresponding classification and segmentation tasks. Such projection-based strategy, by its nature, cannot fully preserve the rich geometric information of the point cloud, and hence it is far from the ideal solution. PointNet~\cite{qi2017pointnet} is the pioneering DNN model that can directly consume the raw point cloud input. By introducing a simple symmetric function to accumulate the features, PointNet preserves the permutation invariance of point cloud data very efficiently. Since then, many architecture-level innovations, such as PointNet++~\cite{qi2017pointnet++} and DGCNN~\cite{wang2019dynamic}, have been proposed to efficiently exploit and capture the local geometric structure. These recent progress further brings the state-of-the-art performance in various point cloud processing tasks.

\textbf{Noise Injection-based Regularization for Image-domain DNN.} Noise injection-based regularization has been widely used in image-domain DNN training to alleviate the overfitting problem. Dropout~\cite{srivastava2014dropout}, as the first work that drops some information/features during the training procedure, successfully demonstrates the huge benefits of this methodology. Since then, many follow-up variants, including but not limited to Droppath~\cite{cai2019effective}, DropBlock~\cite{ghiasi2018dropblock}, ZoneOut~\cite{krueger2016zoneout}, CutOut~\cite{devries2017improved}, Variational Dropout~\cite{kingma2015variational}, have been proposed and applied on the different components of DNN training (e.g., filter channel and feature map) and different DNN model types (e.g., CNN and RNN). According to their extensive experiments, many dropping-related factors, such as rate, schedule and position, have significant impacts on the overall regularization performance.

\textbf{Regularization in Point Cloud Processing.} Unlike their counterparts in image processing, regularization techniques are rarely studied for point cloud processing. To date, most of existing point cloud-oriented regularization works focus on data augmentation. Two most recent progress along this direction are PointMixUp~\cite{10.1007/978-3-030-58580-8_20} and RSMix~\cite{lee2021regularization}, which propose to generate new virtual examples via structure-preserving linear interpolation to enhance model generality. On the other side, noise injection-based regularization, such as dropping certain information and features during the training procedure, is even less exploited in the point cloud domain. Though Dropout has been commonly used in the modern point cloud-domain DNNs (e.g., PointNet, PointNet++ and DGCNN), it is just a straightforward adoption in the last MLP of those models; while injecting the noise to the much more important point feature maps generated by the front-end processing layers, to the best of our knowledge, is not investigated before.

\section{Background and Preliminaries}
\label{sec:background}
\subsection{Basics of DNNs for Point Cloud Processing}

\textbf{Point Cloud.} A point cloud, as a set of $n$ unordered points, can be represented as $\mathcal{X} = \{ \mathbf{x}_i \in \mathbb{R}^{3+d}\}_{i=1}^{n}$, where each point $\mathbf{x}_i$ = $(\mathbf{p}_i, \mathbf{q}_i)$ contains both geometric position $\mathbf{p}_i$ and feature information $\mathbf{q}_i$. To be specific, $\mathbf{p}_i \in \mathbb{R}^3$ is the 3D coordinates, and $\mathbf{q}_i \in \mathbb{R}^d$ represents the corresponding $d$-dimensional feature. Notice that the values of $d$ may vary for different data formats. For instance, for the plain black and the RGB-based colorful point cloud data, $d$ is set as 0 and 3, respectively.

\textbf{Processing Points: Independence Style.} According to different network architectures, point cloud data can be processed in the DNNs with different styles. PointNet~\cite{qi2017pointnet} proposes to first operate on each point independently to achieve the important permutation invariance, and then use max pooling and MLP to aggregate all the extracted individual point features. In general, this type of independent processing style is essentially a set function $f: \mathcal{X} \rightarrow \mathbb{R}^{k}$ that maps a point set to a vector as:
\begin{equation}
    f(\mathbf{x}_1, \mathbf{x}_2, ..., \mathbf{x}_n) = \mathop{\mathbf{MLP}} ( \mathop{\mathbf{max}}_{i=1,...,n}{(h(\mathbf{x}_i)})),
\label{eq:setfunction}
\end{equation}
where  $h(\cdot)$ is the overall mapping function of the front-end layers of DNN model. After each point has been independently processed by those front-end layers, the corresponding $n$ vectorized outputs are aggregated by a max pooling function $\mathbf{max(\cdot)}$ to form a vector-format global feature. This global feature is then sent to the back-end multilayer perceptron (MLP) to learn the desired $k$-dimensional global point cloud signature (see Figure~\ref{fig:framework}).

\textbf{Processing Points: Aggregation Style.} As indicated in~\cite{qi2017pointnet, qi2017pointnet++}, processing each point independently largely neglects the geometric relationship among different points. Therefore, the state-of-the-art point cloud-domain DNNs, e.g., DGCNN and PointCNN~\cite{wang2019dynamic, li2018pointcnn}, adopt to process and aggregate the information of each point and its neighbors in an explicit way. Consequently, the above described set function $f: \mathcal{X} \rightarrow \mathbb{R}^{k}$ can be further generalized as:
\useshortskip
\begin{gather}
    f(\mathbf{x}_1, \mathbf{x}_2, ..., \mathbf{x}_n) = \mathop{\mathbf{MLP}} (\mathop{\mathbf{max}}_{i=1,...,n} (\mathbf{X}^{(L)}_{i,:}) ), \nonumber \\ 
    \mathbf{X}^{(l+1)} =  h^{(l)}(\mathbf{X}^{(l)}), 0 \leq l \leq L-1.
\label{eq:global}
\end{gather}

Here we assume the model requires $L$ times of point feature updates before the max pooling operation. For the $l$-th point feature update, $h^{(l)}: \mathbb{R}^{n^{(l)} \times m^{(l)}} \rightarrow \mathbb{R}^{n^{(l+1)} \times m^{(l+1)}}$ is the update function, and $\mathbf{X}^{(l)} \in \mathbb{R}^{n^{(l)} \times m^{(l)}}$ is the corresponding point feature map for the entire $n^{(l)}$ points associated with this update, where each row $\mathbf{X}^{(l)}_{i,:}$ represents the $m^{(l)}$-dimensional feature for the $i$-th point. For the raw point cloud input, $\mathbf{X}^{(0)} = [\mathbf{x}_1, \mathbf{x}_2, ..., \mathbf{x}_n]^T$ with $n^{(0)} = n$ and $m^{(0)}=d+3$ (see Figure~\ref{fig:framework}).

\subsection{Existing Point Cloud-domain Dropout}

As mentioned in Section \ref{sec:intro}, noise injection-based regularization is rarely studied in the point cloud domain. Currently this strategy is only applied on the two simple stages of the entire point-cloud processing pipeline.

\textbf{Dropout on the Back-end MLP.}
The state-of-the-art point cloud-domain DNNs are typically equipped with an back-end MLP to learn the desired global point cloud signature from the global point feature (see Figure~\ref{fig:framework}). Consider the most conventional use of Dropout in the image domain is just on the MLP; most point cloud-domain DNNs naturally also perform Dropout operation on their back-end MLPs. Hence the optimization objective of the entire DNN, e.g., for shape classification task, can be formulated as: 
\useshortskip
\begin{equation}
    \mathop{min}\mathbb{E}_\mathbf{z} \left\Vert \hat{\mathbf{y}} - \mathop{\mathbf{MLP}}( \mathbf{diag}(\mathbf{z}) (\mathop{\mathbf{max}}_{i=1,...,n} (\mathbf{X}^{(L)}_{i,:}) )) \right\Vert^2_F, 
\label{eq:dropout}
\end{equation}
where $\hat{\mathbf{y}}$ is the vectorized ground truth and $\left\Vert \cdot \right\Vert^2_F$ denotes squared Frobenius norm. $\mathbf{diag(z)}$ is an $m^{(L)} \times m^{(L)}$ diagonal matrix with $\mathbf{z}$ as the diagonal. $\mathbf{z}_{i} \sim Ber(1-\theta)$, as the $i$-th entry of $\mathbf{z}$, is \textit{i.i.d.} Bernoulli with dropping rate $\theta$.

\textbf{Dropout on the Raw Input Points.} In PointNet++~\cite{qi2017pointnet}, Dropout is also performed on the raw input point cloud data. As illustrated in Figure~\ref{fig:framework}, the input points are randomly dropped with various densities in different training epochs. From the perspective of regularization taxonomy, such random dropping operation on the input data is essentially a type of data augmentation method. Hence the optimization objective of the DNN (e.g., for shape classification task) is:
\useshortskip
\begin{gather}
\mathop{min}\mathbb{E}_\mathbf{z}\left\Vert \hat{\mathbf{y}} - \mathop{\mathbf{MLP}} (\mathop{\rm{max}}_{i=1,...,n}( \mathbf{X}^{(L)}_{i,:}))  \right\Vert^2_F, \nonumber \\
\mathbf{X}^{(L)} = h^{(L-1)} \cdots h^{(1)}(h^{(0)}(\mathbf{diag}(\mathbf{z}) \mathbf{X}^{(0)})),
\label{eq:droppoint}
\end{gather}
where $\mathbf{diag}(\mathbf{z})$ is an $n^{(0)} \times n^{(0)}$ diagonal matrix with $\mathbf{z}_{i} \sim Ber(1-\theta)$. $\mathbf{X}^{(0)}$ is raw point cloud inputs, and diagonal matrix $\mathbf{diag}(\mathbf{z})$ randomly zeros a subset of the input points.

\section{Our Methods}
\label{sec:methods}

\subsection{Motivation}   
As formulated in Eq.~\ref{eq:dropout} and~\ref{eq:droppoint}, the current point cloud-domain dropping operations only perform the Dropout either in the very early stage (e.g., on the raw input data) or in the very late stage (e.g., on the back-end last MLP) of the entire processing pipeline. However, these two existing approaches are actually not involved with the main body of DNNs -- the front-end processing layers such as shared MLPs in PointNet and EdgeConv in DGCNN. From the perspective of feature learning, these front-end layers play the critical role for high-performance point cloud processing: they are in charge of preserving, extracting and learning the important multi-level point features in both Euclidean and semantic space. Neglecting this precious regularization opportunity, evidently, will severely limit the performance of DNN models in the point cloud domain.

Motivated by this observation, we propose to systematically investigate the noise injection-based regularization strategy on the important point features. To be specific, we propose three different regularization techniques, namely \textbf{DropFeat}, \textbf{DropPoint} and \textbf{DropCloud}, aiming to inject noise at feature, point and cloud level, respectively.

\subsection{DropFeat: Drop the Features of Points}
In the image domain, Dropout is performed on the activation map with randomly dropping at the pixel level. Following the similar principle, we propose to randomly drop some feature from the entire point feature map $\mathbf{X}^{(L)}$. This feature-level dropping strategy, namely \textit{DropFeat}, is illustrated in  Figure~\ref{fig:framework}. To be specific, for an $n^{(l)}\times m^{(l)}$ point feature map $\mathbf{X}^{(L)}$ where each row represents $m^{(l)}$-dimensional feature belongs to one individual point, DropFeat randomly zeros out some entries of $\mathbf{X}^{(L)}$. In other words, partial (instead of the entire) feature information of the partial (instead of the entire) points are removed. In this scenario, the global distribution of the entire point sets are still preserved, and thereby increasing the generality of the trained models without sacrificing task performance. Hence the optimization objective of the DNN models, e.g., for shape classification task, can be formulated as follows:
\useshortskip
\begin{gather}
\mathop{min}\mathbb{E}_\mathbf{Z}\left\Vert \hat{\mathbf{y}} - \mathop{\mathbf{MLP}} (\mathop{\mathbf{max}}_{i=1,...,n}( \mathbf{X}^{(L)}_{i,:}))  \right\Vert^2_F, \nonumber \\
\mathbf{X}^{(L)} = \mathbf{Z}^{(L-1)} \odot h^{(L-1)} \cdots 
h^{(1)}(\mathbf{Z}^{(1)} \odot h^{(0)}(\mathbf{X}^{(0)})), \nonumber \\
where\; \mathbf{Z}_{ij}^{(l)} \sim Ber(1-\theta).
\end{gather}
Here $\odot$ is the Hardmard product and $\mathbf{Z}^{(l)}(0<l<L)$ is the dropping mask. $\mathbf{Z}^{(l)}$ is placed after the $(l-1)$-th point feature update, and $\theta$ is the pre-set dropping rate. 

\subsection{DropPoint: Drop the Points with the Features}
As described above, DropFeat only drops part of feature information for part of the points during point feature update. From the perspective of regularization, this type of noise injection is relatively conservative. Recall that in PointNet++, some raw input points, which are associated with $(d+3)$-dimensional features, can be entirely dropped in a random way. Such point-wise dropping, surprisingly, can further enhance the generality and robustness of the trained models. In fact, as long as the dropping is performed in a uniform way, such point-wise dropping operation will only make the point sets uniformly sparser -- the critical local and global geometric information can still be preserved, thereby retaining or even improving model performance. 

Inspired by this phenomenon, we propose \textit{DropPoint}, a regularization strategy that performs random point-level dropping operation on the point features. Figure~\ref{fig:framework} illustrates the key idea of DropPoint. For an $n^{(l)}\times m^{(l)}$ point feature map $\mathbf{X}^{(L)}$, DropPoint randomly zeros out some entire rows, where each row  represents one individual point containing $m^{(l)}$-dimensional feature. In other words, all the associated feature information belonging to the randomly selected points are dropped. In general, the optimization objective of the DropPoint-regularized DNN models, e.g., for shape classification task, can be formulated as follows:
\useshortskip
\begin{gather}
\mathop{min}\mathbb{E}_\mathbf{z}\left\Vert \hat{\mathbf{y}} - \mathop{\mathbf{MLP}} (\mathop{\mathbf{max}}_{i=1,...,n}( \mathbf{X}^{(L)}_{i,:}))  \right\Vert^2_F,  \nonumber \\
\mathbf{X}^{(L)} = \mathbf{diag}(\mathbf{z}^{(L-1)})  h^{(L-1)} \cdots h^{(1)} \nonumber (\mathbf{diag}(\mathbf{z}^{(1)})  h^{(0)}(\mathbf{X}^{(0)})), \nonumber\\
where \; \mathbf{z}^{(l)}_i \sim Ber(1-\theta) .
\end{gather}
Here $\mathbf{diag}(\mathbf{z}^{(l)})(0 < l < L)$ is the $n^{(l)} \times n^{(l)}$ diagonal matrix spanned by $\mathbf{z^{(l)}}$, while $\theta$ is the pre-set dropping rate. 

\subsection{DropCluster: Drop the Clusters of Points}
\label{subsec:dropcluster}

Beyond the feature-level dropping (DropFeat) and point-level dropping (DropPoint), we further explore the possibility of more aggressive dropping strategy at higher level. To be specific, we propose \textit{DropCluster}, a technique that performs random dropping operation on the clusters of the neighboring points. Historically, such neighborhood-aware dropping strategy is originated from DropBlock~\cite{ghiasi2018dropblock}, which bounds and drops the contiguous regions of the feature map in the image domain. As~\cite{ghiasi2018dropblock} indicates, the contiguous region-free dropping (e.g., Dropout) is not effective in removing semantic information due to the spatial correlation among the nearby activations. Instead, dropping the contiguous region of feature map can effectively remove the semantic information in the correlated area, and hence the models are forced to learn more representative features. 

\begin{figure}
    \centering
    \includegraphics[width =\linewidth]{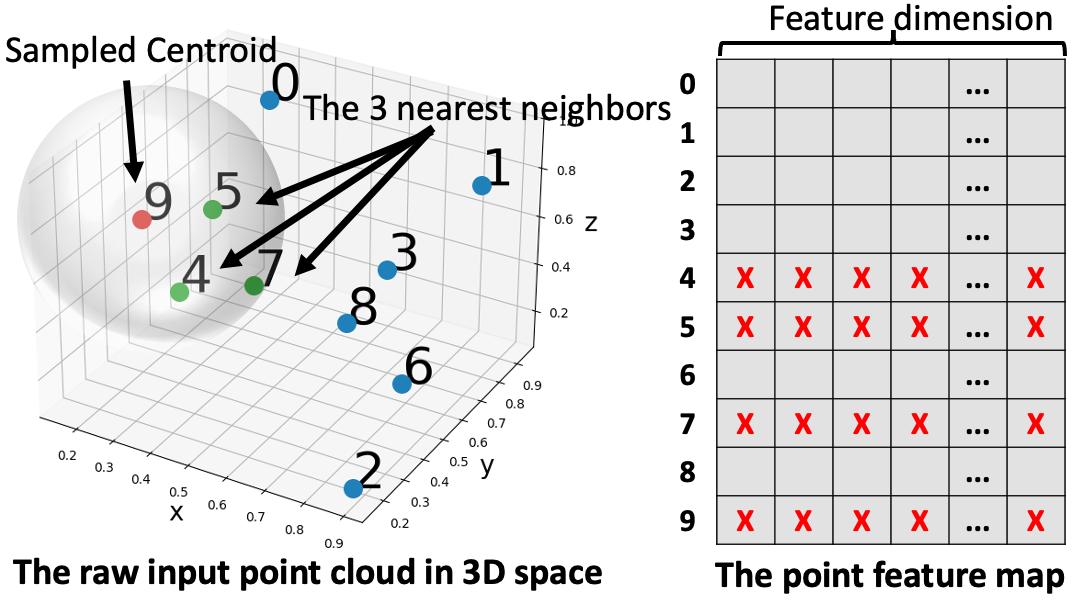}
    \caption{An example of DropCluster. Point 9 is the sampled centroid. Three nearest neighboring points (4, 5 and 7) are selected when cluster size $\gamma=4$. \textbf{The nearest neighbors are determined by the distances measured by the 3D ordinates of input point cloud, instead of the positions in the current point feature map.}}
    \label{fig:cluster}
    \vspace{-5mm}
\end{figure}

Inspired by DropBlock, we believe dropping the neighboring points in the point cloud domain can bring the similar benefits: the nearby points typically contain the closely related information, hence a neighborhood-aware dropping strategy can effectively remove certain geometric and semantic information, thereby pushing the DNN models to enhance its capability for feature learning. Figure \ref{fig:cluster} illustrates the main mechanism of DropCluster. Here some points in the point feature map are first randomly selected. With each of those points as the centroid, multiple clusters of points, which contain the selected points and their neighboring points, are then dropped during the training.
 
As a neighborhood-aware dropping strategy, the performance of DropCluster highly depends on two factors: \textit{distance calculation} and \textit{neighboring point selection}. Considering the raw point cloud input data is defined on the Euclidean space, DropCluster directly utilizes the 3D coordinates of input points to calculate spatial distances. Notice that for some DNN models such as DGCNN, this computation can be even saved since the distance calculation has already been done in their K-Nearest Neighbors (KNN) step.

\textbf{Neighboring Point Selection.} To determine and adjust the size of the dropped region, DropCluster introduces a hyperparameter $\gamma$ to determine the number of the selected points in one cluster. To be specific, once a point is randomly selected, its nearest $\gamma-1$ points, together with their associated features, will be dropped as well. Notice that since $\gamma$ points are now dropped simultaneously for each cluster, the sampling rate for the centroid points becomes $1-\theta/\gamma$ instead of $1-\theta$. In general, the optimization objective of the DropCluster-regularized DNN models, e.g., for shape classification task, can be formulated as follows:
\useshortskip
\begin{gather}
\mathop{min}\mathbb{E}_\mathbf{z}\left\Vert \hat{\mathbf{y}} - \mathop{\mathbf{MLP}} (\mathop{\mathbf{max}}_{i=1,...,n}( \mathbf{X}^{(L)}_{i,:}))  \right\Vert^2_F, \nonumber\\
\mathbf{X}^{(L)} = \mathbf{diag}(\mathbf{z}^{(L-1)}) h^{(L-1)}  \cdots h^{(1)}(\mathbf{diag}(\mathbf{z}^{(1)}) h^{(0)}(\mathbf{X}^{(0)})), \nonumber \\
\mathbf{z}_i^{(l)} = 0,\, if\,i\in \{i \,|\, \exists \mathbf{w}_j^{(l)} =0,\, i \in \{j \cup \mathbf{knn}(j, \gamma - 1 )\}\}, \nonumber \\
where\, \mathbf{w}_j^{(l)} \sim Ber(1 - \theta / \gamma ).
\end{gather}
Here $\mathbf{diag}(\mathbf{z}^{(l)})$ is an $n^{(l)} \times n^{(l)}$ diagonal matrix spanned by $\mathbf{z^{(l)}}$, and $\mathbf{w}^{(l)}$ is an $n^{(l)}$-length binary vector to record the index of the sampled centorid point. Each entry of $\mathbf{w}^{(l)}$ is sampled from the Bernoulli distribution with the probability $1 - \theta / \gamma$, where 0 denotes the corresponding entry is the sampled centroid. The function $\mathbf{knn}(j, \gamma - 1)$ returns the set of $\gamma - 1$ nearest neighbors' indices of the point $j$. Notice that when $\gamma=1$, DropCluster converges to DropPoint.

\section{Experiments}
\subsection{Experimental Setup}
\textbf{Dataset.} We evaluate our methods (DropFeat, DropPoint and DropCluster) on three 3D point cloud datasets ModelNet40~\cite{wu20153d}, ShapeNet part dataset~\cite{yi2016scalable} and Stanford Large-Scale 3D Indoor Spaces Dataset (S3DIS)~\cite{armeni20163d} for 
shape classification, part segmentation and semantic segmentation tasks, respectively. To be specific,  For \textbf{ModelNet40}, we use 9843 models for training and 2468 models for testing. In each model 1024 points are sampled and rescaled into the unit sphere. For \textbf{ShapeNet}, we use 14006 shapes for training and 2874 shapes for testing. In each shape 2048 points are sampled and at most labeled with five parts. For \textbf{S3DIS},  we follow the same 6-fold strategy used in PointNet. For each example in S3DIS, 4096 points are sampled and each input point is represented as a 9D feature vector (XYZ, RGB and normalized spatial coordinates).

\textbf{Network Architecture.} We select three popular backbone networks, PointNet, PointNet++ with multi-scale groping (MSG) and DGCNN, to demonstrate the generality of our proposed noise injection-based regularization approaches. All of these three networks are evaluated on ModelNet40, ShapeNet and S3DIS datasets.

\textbf{Implementation Details.} We conduct our experiments on Nvidia TITAN RTX GPUs using PyTorch framework. For all the evaluations for PointNet and PointNet++, we use the ADAM optimizer with initial learning rate as 0.001. For the evaluations for DGCNN, we adopt SGD optimizer with initial learning rate as 0.1 and the momentum as 0.9. The batch size is set from 16 to 32 across different tasks.

\subsection{Ablation Study for $\mathbf{\theta}$, $\mathbf{\gamma}$ and Dropping Positions}
\label{sec:ablation}
As analyzed in Section \ref{sec:methods}, the performance of our proposed noise injection-based regularization approaches is highly determined by three key factors: the dropping rate $\theta$, the place where to apply the dropping, and cluster size $\gamma$ (for DropCluster). To study their individual impact on the DNN performance, we perform ablation study on an example experimental setting: Applying DropCluster on DGCNN for shape classification task (ModelNet40 dataset).

\begin{figure}[h]
\vspace{-4mm}
    \centering
    \subfigure[The drop rate $\theta$.]{
    \label{fig:rate}
    \includegraphics[width=1.55in]{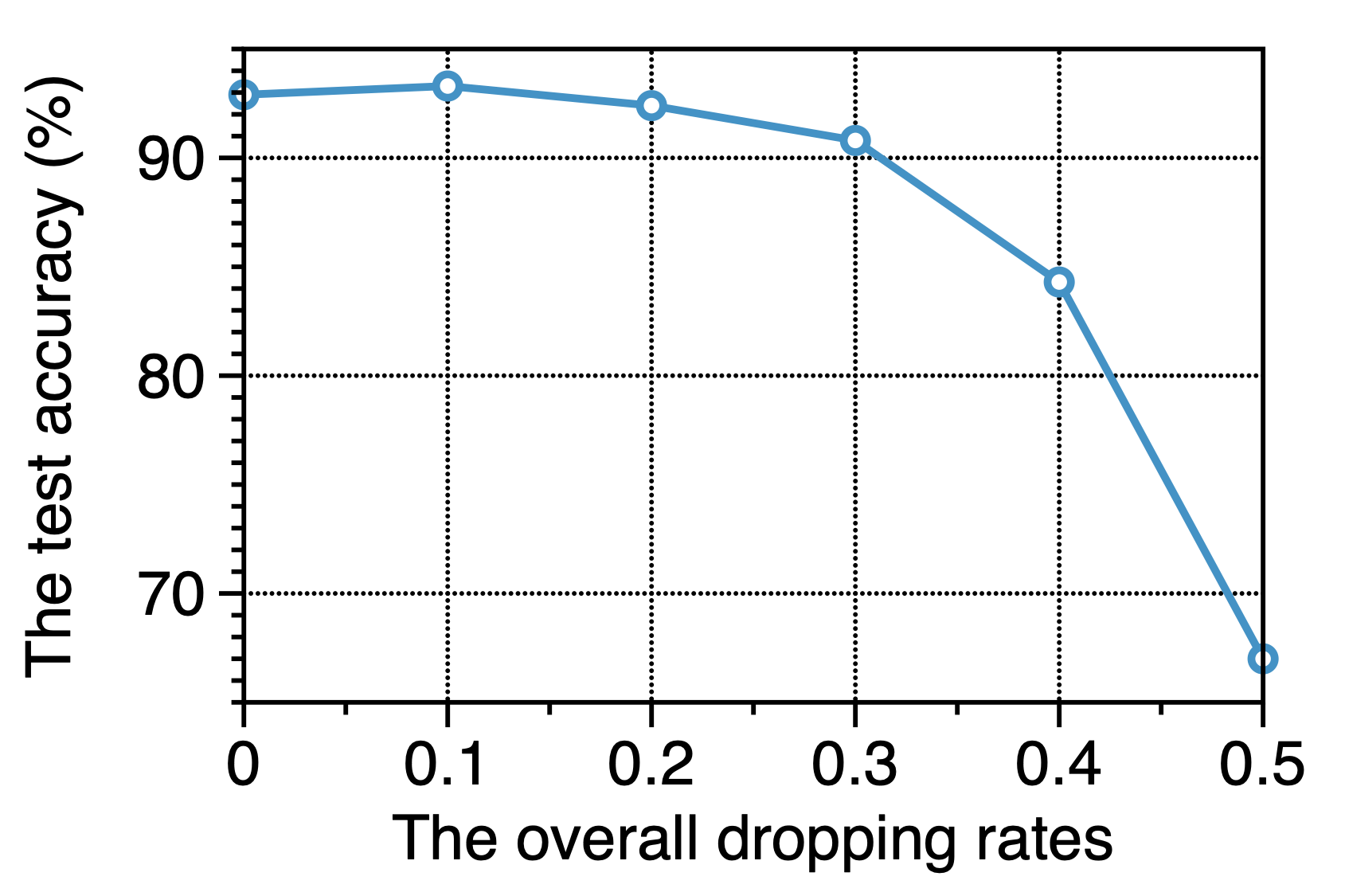}}
    \subfigure[The cluster size $\gamma$.]{
    \label{fig:size}
    \includegraphics[width=1.55in]{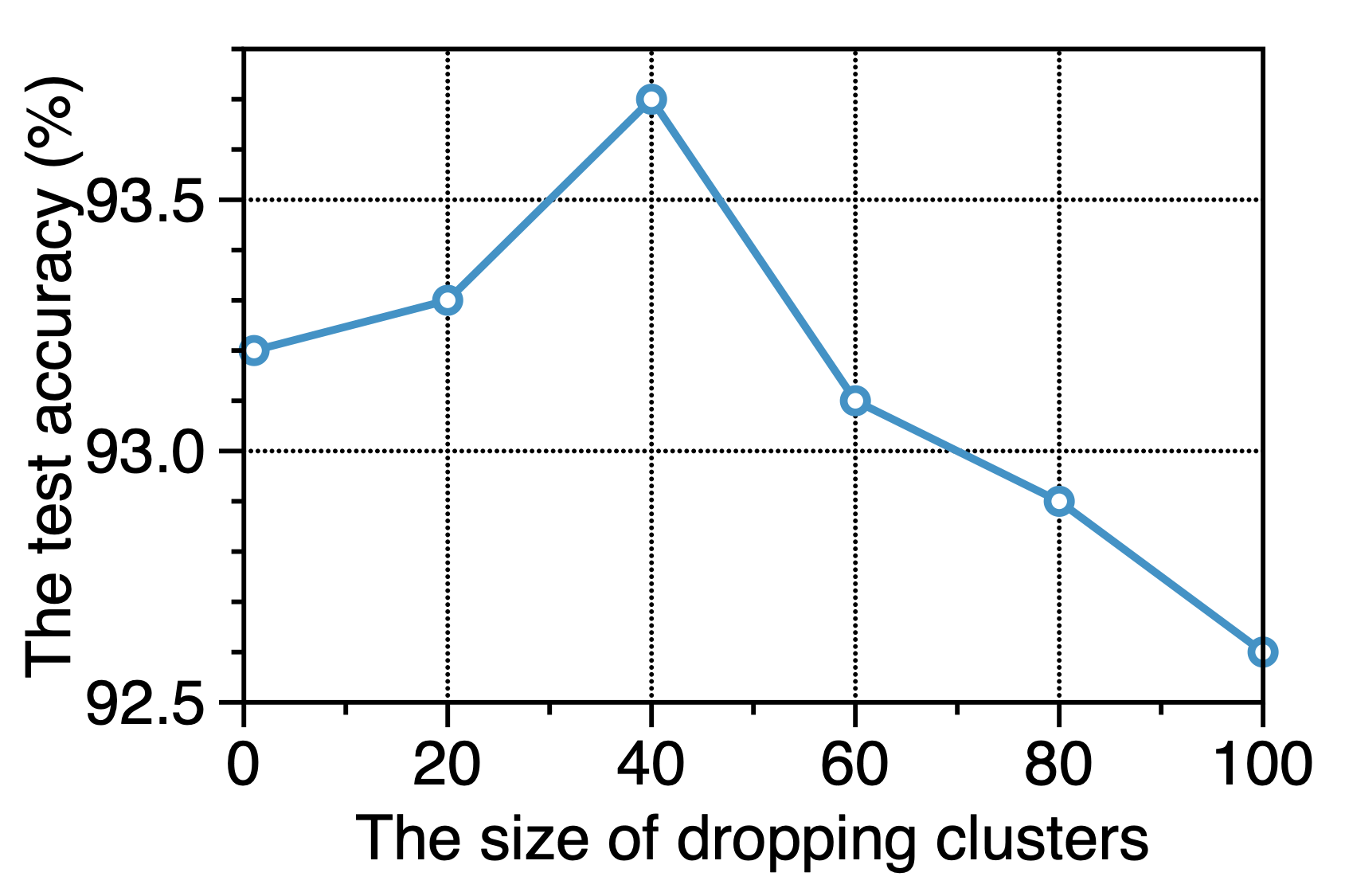}}
    \caption{Ablation study for dropping rate and cluster size (DropCluster-regularized DGCNN on ModelNet40).  (a) Different dropping rate $\theta$ with the fixed the cluster size $\gamma=20$. (b) Different different values of $\gamma$ with the fixed dropping rate $\theta = 0.1$. DropCluster is performed after the 1st and 2nd EdgeConv blocks.}
    \label{fig:ablation_rate_size}
    \vspace{-3mm}
\end{figure}

\textbf{The Dropping Rate $\theta$.} 
We first investigate the impact of overall dropping rate $\theta$. Figure~\ref{fig:rate} shows the test accuracy of DropCluster-regularized DGCNN on ModelNet40 with respect to different values of $\theta$. Here the cluster size $\gamma$ is set as 20. It is seen that when $\theta=0.1$ the test accuracy reaches its maximum ($0.3\%$ higher than non-dropping case), and then it quickly decreases with higher dropping rates. This phenomenon indicates that, unlike the case in the image domain (e.g. dropping rate=0.5 for Dropout), smaller dropping rate is preferred in the point cloud domain. We hypothesize that this trend may be caused by the relatively smaller number of points in the point cloud data -- typically the point feature map only has $1024\sim 4096$ points; while the feature map in image domain can have tens of thousand activation values. So too aggressive dropping strategy may negatively affect the overall performance.

\textbf{The Cluster Size $\gamma$.} Figure~\ref{fig:size} shows the overall test accuracy of DGCNN with respect to different cluster sizes. Here the overall dropping rate $\theta$ is set as $0.1$. It is seen that the test accuracy significantly increases when $\gamma=20$, peaks when $\gamma=40$, and then decreases with larger $\gamma$. This phenomenon indicates that a proper setting of $\gamma$ in the medium range is very important to achieve a "sweet point" -- dropping either too few or too many neighboring points would impair the capability of learning geometric information. Also notice that the accuracy with $\gamma=1$, which just means using DropPoint, is lower than many $\gamma\neq1$ cases. This verifies our prior hypothesise that dropping nearby points together with proper dropping rates is more effective.

\textbf{Where to Drop.} We further study where to apply Dropping operation on the DNN models can achieve best performance. As shown in Table~\ref{tab:layer}, we apply DropCluster at different positions of DGCNN. Here because the DGCNN classification model has 4 EdgeConv blocks, namely EdgeConv 1,2,3,4 in Table~\ref{tab:layer}, we evaluate different combinations when applying DropCluster on the point feature maps output from those EdgeConv blocks. From this table we can find that it is better to apply dropping operation at the earlier stages rather than later stages of DNN models. We hypothesize that this is because the point features extracted in the earlier stages of DNN intend to represent lower-level feature that is more noise tolerate; while the point features extracted in the later stages are more high-level condensed features that are sensitive to noise injection.

\begin{table}[h]
    \centering
    \begin{tabular}{c|c|c|c|c}
    \toprule
         \makecell{Edge \\ Conv1} & \makecell{Edge \\\ Conv2} & \makecell{Edge \\\ Conv3} & \makecell{Edge \\\ Conv4} & \makecell{Accuracy \\ (\%)}  \\
         \hline \hline
         & & & & 92.9 (Baseline) \\
         $\checkmark$ & & & & 93.3 (0.4$\uparrow$)\\
         & $\checkmark$ & & & 93.4 (0.5$\uparrow$) \\
         & & $\checkmark$ & & 92.8 (0.1$\downarrow$) \\
         & & & $\checkmark$ & 92.9 (0.0-) \\
         $\checkmark$ & $\checkmark$ & & & \textbf{93.7 (0.8$\uparrow)$} \\
         $\checkmark$ & $\checkmark$ & $\checkmark$ & & 92.8 (0.1$\downarrow$) \\
         $\checkmark$ & $\checkmark$ & $\checkmark$ & $\checkmark$ & 92.8(0.1$\downarrow$) \\
         & & $\checkmark$ & $\checkmark$ & 93.0 (0.1$\uparrow$) \\
    \bottomrule
    \end{tabular}
    \caption{Ablation study for dropping positions (DropCluster-regularized DGCNN on ModelNet40 dataset). \textbf{If an EdgeConv block is selected, it means to perform DropCluster on the point feature map output from that block.} Here $\theta=0.1$ and $\gamma=40$.}
    \label{tab:layer}
    \vspace{-2mm}
\end{table}

\begin{table*}[]
\centering
\begin{tabular}{lcccccc}
\toprule
   \multirow{2}{*}{\makecell{Backbone \\ Architecture}} & \multirow{2}{*}{\makecell{Baseline \\ Accuracy ($\%$)}} & \multicolumn{2}{c}{Data Augment-based Regularization} & \multicolumn{3}{c}{Noise Injection-based Regularization}\\
    \cmidrule(r){3-4}
    \cmidrule(r){5-7}
  &  & RSMix~\cite{lee2021regularization} & PointMixup~\cite{10.1007/978-3-030-58580-8_20} & DropFeat & DropPoint & DropCluster \\
\midrule
PointNet  & 89.2 & 89.4(0.2$\uparrow$) & 89.9(0.7$\uparrow$) & 89.8(0.6$\uparrow$) & 90.4(1.2$\uparrow$) & \textbf{90.7(1.5$\uparrow$)} \\
\midrule
PointNet++ & 91.9 & 93.0(1.1$\uparrow$)  & 92.7(0.8$\uparrow$) & 92.7(0.8$\uparrow$) & 93.0(1.1$\uparrow$) & \textbf{93.2(1.3$\uparrow$)} \\
\midrule
DGCNN & 92.9 & 93.5(0.6$\uparrow$) & 93.1(0.2$\uparrow$) & 93.0(0.1$\uparrow$) & 93.2(0.3$\uparrow$) & \textbf{93.7(0.8$\uparrow$)} \\
\bottomrule
\end{tabular}
\caption{Overall accuracy of different regularization approaches on ModelNet40 shape classification dataset.}
\label{tab:comp}
\end{table*}

\begin{table*}[h]
\centering
\small
\setlength{\tabcolsep}{3pt}
\begin{tabular}{l|c|cccccccccccccccc}
\toprule
& \textbf{\makecell{Mean \\ IoU}} & Aero & Bag & Cap & Car & Chair & \makecell{Ear \\ Phone} & Guitar & Knife & Lamp & Laptop & Motor & Mug & Pistol & Rocket & \makecell{Skate \\ Board} & Table \\
\midrule
\# Shapes & & $2690$ & $76$ & $55$ & $898$ & $3758$ & $69$ & $787$ & $392$ & $1547$ & $451$ & $202$ & $184$ & $283$ & $66$ & $152$ & $5271$ \\
\midrule
PointNet & $83.7$ & $83.4$ & $78.7$ & $82.5$ & $74.9$ & $89.6$ & $73.0$ & $91.5$ & $85.9$ & $80.8$ & $95.3$ & $65.2$ & $93.0$ & $81.2$ & $57.9$ & $72.8$ & $80.6$ \\
+DropFeat & $84.1$ & $81.7$ & $73.9$ & $81.0$ & $73.6$ & $90.4$ & $74.1$ & $90.3$ & $85.0$ & $80.5$ & $95.2$ & $59.4$ & $90.9$ & $79.1$ & $55.5$ & $73.5$ & $82.2$ \\
+DropPoint & $83.9$ & $81.6$ & $76.9$ & $82.0$ & $73.8$ & $90.2$ & $72.1$ & $90.3$ & $86.8$ & $80.4$ & $95.3$ & $57.9$ & $92.3$ & $77.3$ & $53.8$ & $72.0$ & $82.1$ \\
+DropCluster & \textbf{84.2} & $82.4$ & $76.6$ & $88.5$ & $74.5$ & $90.2$ & $74.2$ & $90.8$ & $87.0$ & $81.1$ & $95.5$ & $61.9$ & $94.1$ & $79.5$ & $52.6$ & $73.4$ & $82.4$ \\
\midrule
PointNet++ & $85.1$ & $82.4$ & $79.0$ & $87.7$ & $77.3$ & $90.8$ & $71.8$ & $91.0$ & $85.9$ & $83.7$ & $95.3$ & $71.6$ & $94.1$ & $81.3$ & $58.7$ & $76.4$ & $82.6$ \\
+DropFeat & $85.5$ & $83.0$ & $81.9$ & $89.2$ & $78.3$ & $90.6$ & $74.3$ & $91.2$ & $86.4$ & $83.7$ & $95.7$ & $74.5$ & $95.5$ & $81.8$ & $61.3$ & $76.6$ & $82.6$ \\
+DropPoint & \textbf{85.7} & $83.4$ & $81.0$ & $87.6$ & $78.1$ & $91.3$ & $77.6$ & $90.8$ & $87.9$ & $84.2$ & $95.6$ & $73.2$ & $95.2$ & $82.2$ & $60.8$ & $75.5$ & $83.3$ \\
+DropCluster & $85.6$ & $83.1$ & $80.6$ & $85.7$ & $78.3$ & $90.7$ & $75.8$ & $91.3$ & $87.5$ & $84.5$ & $95.6$ & $70.5$ & $93.8$ & $81.9$ & $60.4$ & $76.7$ & $83.5$ \\
\midrule
DGCNN & $85.2$ & $84.0$ & $83.4$ & $86.7$ & $77.8$ & $90.6$ & $74.7$ & $91.2$ & $87.5$ & $82.8$ & $95.7$ & $66.3$ & $94.9$ & $81.1$ & $63.5$ & $74.5$ & $82.6$ \\ 
+DropFeat & $85.3$ & $84.2$ & $82.6$ & $83.9$ & $77.2$ & $90.8$ & $74.5$ & $90.9$ & $87.7$ & $83.1$ & $95.4$ & $64.7$ & $94.2$ & $81.3$ & $56.6$ & $74.3$ & $83.4$ \\
+DropPoint & $85.3$ & $84.1$ & $82.8$ & $84.1$ & $78.2$ & $91.4$ & $74.6$ & $90.4$ & $85.3$ & $83.5$ & $94.9$ & $64.6$ & $94.1$ & $81.6$ & $572.$ & $74.4$ & $83.3$ \\
+DropCluster & \textbf{85.4} & $84.5$ & $83.1$ & $84.8$ & $77.5$ & $90.9$ & $75.6$ & $91.1$ & $88.5$ & $83.3$ & $95.8$ & $64.1$ & $94.6$ & $81.7$ & $58.4$ & $74.7$ & $83.6$ \\
\bottomrule
\end{tabular}
\caption{Performance of different regularization approaches on ShapeNet part segmentation dataset. The evaluation metric is the mean IoU($\%$). \textbf{Notice that RSMix and PointMixup do not report performance on part segmentation task.}}
\label{tab:partseg}
\vspace{0mm}
\end{table*}

\subsection{Shape Classification on ModelNet40 Dataset}
\textbf{Settings.} Based on the analysis in the ablation study, we set the overall dropping rate $\theta=0.1$ for DropFeat, DropPoint and DropCluster on all the shape classification experiments. Also, we set the cluster size $\gamma=40$ for DropCluster when regularizing DGCNN and PointNet. For PointNet++, since it hierarchically samples a portion of the input point cloud, we set $\gamma=20$ to fit the reduced number of points. Meanwhile, we apply DropFeat, DropPoint and DropCluster on the point feature maps output from the first two EdgeConv layers and the first two set abstraction blocks of DGCNN and PointNet++, respectively. For PointNet, the dropping operation is performed on the point feature maps of the first shared linear layers with 64 output channels.

\begin{figure}[h]
    \centering
    \includegraphics[width=\linewidth]{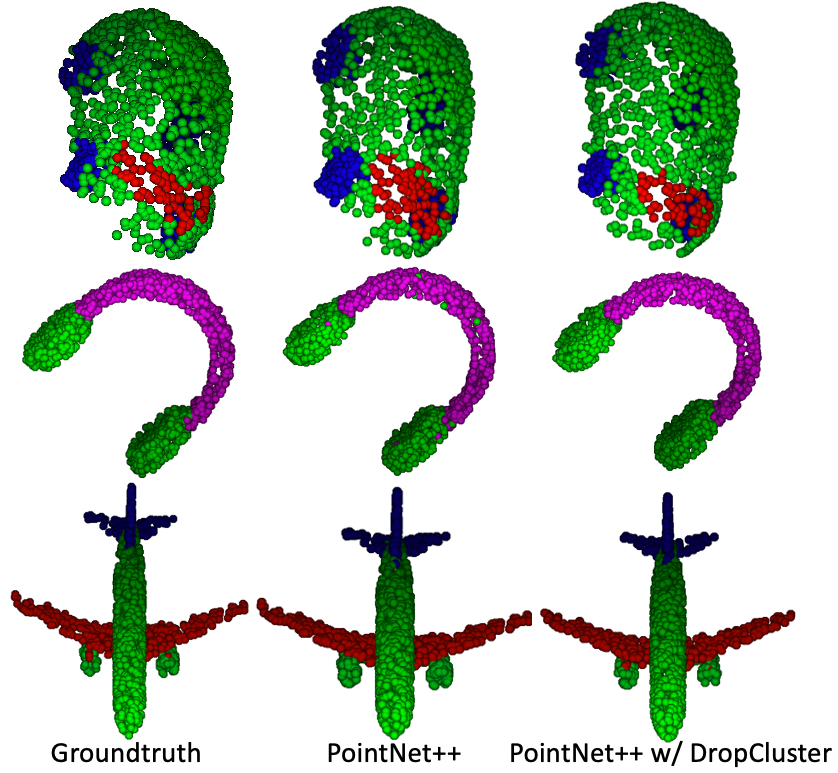}
    \caption{Qualitative results for part segmentation on CAD examples from ShapeNet.}
    \label{fig:partseg}
    \vspace{-3mm}
\end{figure}

\textbf{Results.} Table~\ref{tab:comp} shows the overall accuracy of three backbone networks regularized by different approaches on ModelNet40. Compared with the state-of-the-art data augmentation-based regularization (RSMix~\cite{lee2021regularization} and PointMixup~\cite{10.1007/978-3-030-58580-8_20}), our proposed noise injection-based regularization approaches show the competitive (DropFeat and DropPoint) and better (DropCluster) performance. In particular, DropCluster consistently outperforms RSMix and PointMixup on all the three models. To be specific, DropCluster achieves $1.5\%$, $1.3\%$ and $0.8\%$ overall accuracy improvement on PointNet, PointNet++ and DGCNN, respectively.

\subsection{Part Segmentation on ShapeNet Dataset}

\textbf{Settings.} For part segmentation task we still use the dropping rate $\theta=0.1$ for all the dropping methods on all the backbone networks. Also, the dropping operations are performed at the same positions as described in the shape classification task. Since the inputs of part segmentation models are 2048-point data, we adjust cluster size $\gamma=80$ for PointNet and DGCNN, and $\gamma=40$ for PointNet++. 

\textbf{Results.}
Table~\ref{tab:partseg} shows the performance of the regularized backbone networks on ShapeNet dataset in terms of mean Intersection-over-union (IoU). It is seen that all our proposed three dropping approaches improve the mean IoU for part segmentation task. In particular, DropCluster brings $0.5\%$, $0.5\%$ and $0.2\%$ mean IoU increase for PointNet, PointNet++, and DGCNN, respectively. \textbf{Notice that RSMix and PointMixup do not report performance on part segmentation task.} Also, we show the qualitative part segmentation results in Figure ~\ref{fig:partseg}.

\begin{table}[]
\centering
\begin{tabular}{l|c|c}
\toprule
    \makecell{Backbone \\ Architecture} &  Mean IoU($\%$) & Overall Accuracy($\%$) \\
\hline\hline
    PointNet & 47.7 & 78.6 \\
    w/ DropCluster & \textbf{50.9(3.2$\uparrow$)} & \textbf{81.0(2.4$\uparrow$)} \\
    \midrule
    PointNet++ & 51.3 & 81.7 \\
    w/ DropCluster & \textbf{54.2(2.9$\uparrow$)} &  \textbf{83.9(2.2$\uparrow$)}\\
    \midrule
    DGCNN & 56.1 & 84.1 \\
    w/ DropCluster & \textbf{59.8(3.7$\uparrow$)} & \textbf{85.9(1.8$\uparrow$)} \\
\bottomrule 
\end{tabular}
\caption{Performance of DropCluster on S3DIS semantic segmentation dataset. \textbf{Notice that RSMix and PointMixup do not report performance on semantic segmentation task.}}
\label{tab:semantic}
\vspace{-5mm}
\end{table}

\subsection{Semantic Segmentation on S3DIS Dataset}
\textbf{Setting.} Considering DropCluster consistently outperforms DropFeat and DropPoint in the shape classification and part segmentation tasks, in semantic segmentation task we only evaluate the performance of DropCluster regularization. Here we adopt the same dropping rate ($\theta = 0.1$) and dropping positions that are used in shape classification and part segmentation experiments. In addition we use the same cluster size settings that are used in shape classification ($\gamma=40$ for PointNet and DGCNN and $\gamma=20$ for PointNet++). This is because though each S3DIS input example has large number of point, the average number of points for each semantic object is relatively small -- several objects belonging to different categories exist in one room.

\begin{figure}[h]
    \centering
    \includegraphics[width=\linewidth]{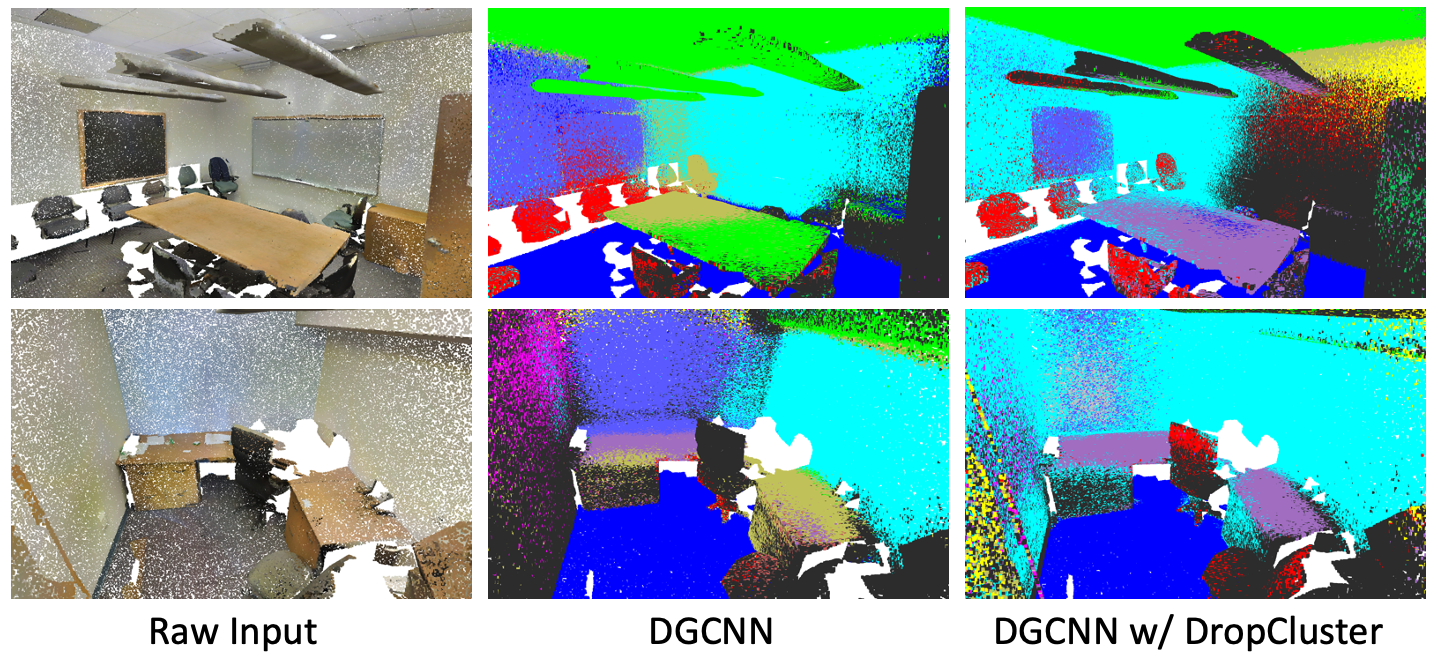}
    \caption{Qualitative results for semantic segmentation.}
    \label{fig:semseg}
    \vspace{-5mm}
\end{figure}

\textbf{Results.} Table~\ref{tab:semantic} shows the performance of DropCluster-regularized backbone networks on S3DIS dataset in terms of both the mean IoU and overall accuracy. It is seen that using DropCluster brings $3.2\%$, $2.9\%$, $3.7\%$ mean IoU improvement on PointNet, PointNet++ and DGCNN, respectively. For overall accuracy, DropCluster enables $2.4\%$, $2.2\%$, $1.8\%$ increase for PointNet, PointNet++ and DGCNN, respectively. \textbf{Notice that RSMix and PointMixup do not report performance on semantic segmentation task.} In addition, We also show the qualitative semantic segmentation results in Figure~\ref{fig:semseg}.

\section{Conclusion}
In this paper, we propose a systematic investigation on noise injection-based regularization for point cloud processing. We develop a series of techniques to inject noise at the different levels of point feature maps of DNN models. Experimental results show our proposed approaches bring significant performance improvement across different DNN models for different point cloud processing tasks.

\newpage
{
\small
\bibliographystyle{ieee_fullname}
\bibliography{main}
}

\end{document}